# A Novel Hybrid Approach for Cephalometric Landmark Detection

MAHSHID MAJD, FARZANEH SHOELEH

*Department of Computer Science and Engineering and Information Technology, Shiraz University, Shiraz, Iran*

*Mollasadra Ave., Shiraz, Iran*

Email: {majd, shoeleh}@cse.shirazu.ac.ir

Abstract. Cephalometric analysis has an important role in dentistry and especially in orthodontics as a treatment planning tool to gauge the size and special relationships of the teeth, jaws and cranium. The first step of using such analyses is localizing some important landmarks known as cephalometric landmarks on craniofacial in x-ray image. The past decade has seen a growing interest in automating this process. In this paper, a novel hybrid approach is proposed for automatic detection of cephalometric landmarks. Here, the landmarks are categorized into three main sets according to their anatomical characteristics and usage in well-known cephalometric analyses. Consequently, to have a reliable and accurate detection system, three methods named edge tracing, weighted template matching, and analysis based estimation are designed, each of which is consistent and well-suited for one category. Edge tracing method is suggested to predict those landmarks which are located on edges. Weighted template matching method is well-suited for landmarks located in an obvious and specific structure which can be extracted or searchable in a given x-ray image. The last but not the least method is named analysis based estimation. This method is based on the fact that in cephalometric analyses the relations between landmarks are used and the locations of some landmarks are never used individually. Therefore the third suggested method has a novelty in estimating the desired relations directly. The effectiveness of the proposed approach is compared with the state of the art methods and the results were promising especially in real world applications.

## 1. INTRODUCTION

Cephalometric analysis is a treatment tool used by dentists and other maxillofacial surgeons. The analysis is done on cephalometric radiographs, namely cephalogram, by identifying some predefined landmarks and measuring the linear and angular relations of them. Several methods of analyses have been used while the best known ones are Steiner, Mc Narama and Downs. Among various landmarks placed on the soft tissues or bones, sixteen landmarks are most widely used in cephalometric analyses, especially these three famous analyses.

For years the landmark localization has been done manually by an expert. The manual landmark identification is the most widely used method while it is a time consuming process and depends highly on the analyzer's expertise. To aid the analyzers, the automatic landmark detection methods have been introduced where the landmarks are localized by intelligent methods. The premier methods were meant just for research but soon after, effective approaches have been presented.

Several approaches based on Image processing and pattern recognition methods have been presented to automate landmark identification. These methods can be classified into following four categories [1]; 1) Knowledge based approaches using image processing techniques, 2) Model based approaches, 3) Soft computing methods, 4) Hybrid models.

Since 1989, the image processing approaches has been widely used to identify the Cephalometric landmarks [2,3]. In these approaches, the image processing techniques, usually edge detection, combined with prior knowledge on typical shape of craniofacial structure are applied to the image regarding to the landmarks locations. Despite the simplicity and availability of these methods, they have two main disadvantages; first of all they are highly dependent on the image quality and secondly they cannot find all kinds of landmarks especially those which are not located on any significant edge. These techniques still has an important role in the automatic detection [4].

Model based approaches include template matching, active shape model and active appearance model [5,6,7]. This category is used for detecting the landmarks located on a visible particular anatomical shape. These methods use this shape to produce a base model and match it through the image. A big advantage of these approaches is their independence from scale rotation and translation while their disadvantage is the need for a model to be matched with the landmark area. Besides, the model deformation must be constrained and it is sensitive to noise.

The most important soft computing technique used in this realm, is Artificial Neural Networks (ANNs) which is an adaptive learning system [8,9,10]. They provide shape variability and noise tolerance. On the other hand, these methods have a training phase where a large number of cephalograms is needed. So, the main drawback of using ANN is the fact that its training phase is time consuming and has high complexity. Support Vector Machines, SVM, is another soft computing technique which is used to detect Cephalometric landmarks [11].

Recently, most researches in this area have been focused to the hybrid approaches [12, 13], where the three above approaches are combined efficiently in order to identify each landmark regarding to its nature more accurately.

In this paper, a new hybrid approach for automatic landmark detection is presented. While image processing detection methods are still the best approach for some landmarks, it is not the answer for all of them. This issue is identical for the other methods. Here, it is tried to take the advantages of the first two approaches and combine them with a totally new idea where the landmarks relations are calculated. To the best of our knowledge, all of the presented approaches were focused on finding the exact location of some landmarks to be used in Cephalometric analyses while the main aim of the most Cephalometric analyses is to find the linear and angular relation of these landmarks. Each of the mentioned approaches in the above four categories is appropriate for a limited number of landmarks and there are still some important landmarks which cannot be accurately localized by any of the presented approaches while they are so important to the Cephalometric analyses.

The presented approach called HALD is tested against a randomly selected dataset of real world cephalometric radiographs. The results show that our method is a practical way for automatic landmark detection and the overall estimation error is acceptable.

The remaining of this paper is organized as follows. Section 2 describes the proposed method in detail. Experiments are given in Section 3 where the method is compared with other well known approaches. The last section includes summary and conclusion.

**2. Proposed Method**

The main aim of this paper is introducing a novel hybrid approach for automatic landmarking of cephalograms using image processing techniques, named HALD (**H**ybrid **A**utomatic **L**andmark **D**etector). As the main cephalometric analyses such as Mc Narama, Steiner and Downs are based on the position of sixteen important landmarks (Table 1), here our system named HALD is designed to detect these landmarks automatically. Figure 1 shows cephalometric radiogram with its landmarks.

Table 1. The searched landmarks.

| Landmarks | Description |
|---|---|
| Menton (Me) | The lowest point on the symphysis of the mandible |
| Pogonion (Pog) | The most anterior point on the contour of the chin |
| Gnathion (Gn) | The most outward and everted point on the profile curvature of the symphysis of the mandible, located midway between pogonion and menton |
| Nasion (N) | The junction of the nasal and frontal bones at the most posterior point on the curvature of the bridge of the nose |
| Sella (S) | The center of the hypophyseal fossa (sella tursica) |
| Anterior Nasal Spine (ANS) | The most anterior point on the maxilla at the nasal base |
| Posterior Nasal Spine (PNS) | The tip of the posterior nasal spine of the palatine bone, at the junction of the soft and hard palate |
| A point (A) | The most anterior point of the maxillary apical base |
| B point (B) | The innermost curvature from chin to alveolar junction |
| Upper Incisor Tip (UIT) | The tip of the crown of the upper central incisor |
| Upper Incisor Apex (UIA) | The root apex of the upper central incisor |
| Porion (Po) | The uppermost point of the external ear meatus |
| Orbitale (Or) | A point midway between the lowest point on the inferior margin of the two orbits |
| Gonion (Go) | A point midway between the points representing the middle of the curvature at the left and right angles of the mandible |
| Lower Incisor Tip (LIT) | The tip of the crown of the lower central incisor |
| Lower Incisor Apex (LIA) | The root apex of the lower central incisor |

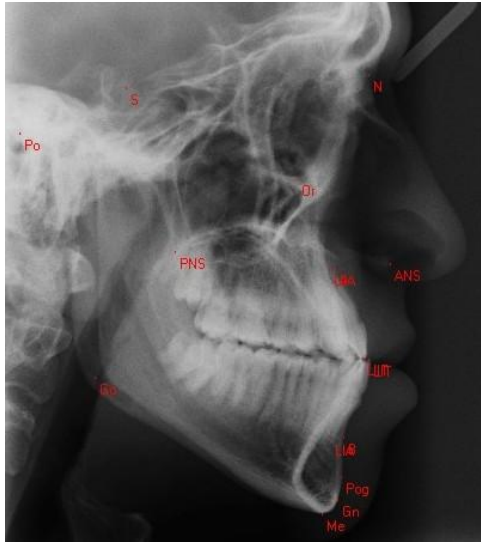

Figure 1. A cephalometric radiogram with its localized landmarks.

According to the cephalograms, each landmark is located on a particular region of skull where soft tissues, nerves and blood vessels in this area can cause unique anatomical characteristics. For example, while Or is surrounded in a pool of soft tissues, ME is on the very clear edge of mandible. Therefore, it might not be wise to use same method to detect these sixteen landmarks. Consequently, it could be useful to use an approach based on hybrid of different methods that each of which is designed to detect a group of landmarks with similar characteristics.

The overall view of our proposed method is pictured in Figure 2. The introduced landmarks can be categorized in three main groups by considering their anatomical characteristics: 1) landmarks which are located on specific edges, 2) landmarks surrounded by a unique structure which can be identified in the cephalograms and 3) landmarks that do not belong to the previous categories. Regarding to this categorization, our approach (HALD) consists of three main mechanisms that each of which is designed to identify and detect the landmarks of one mentioned category. These proposed mechanisms are: 1) Edge tracing method, which is suggested to predict those landmarks located on edges. 2) Weighted template matching method, which is well-suited for landmarks located in an obvious and specific structure which can be extracted or searchable in a given x-ray image. 3) The last but not the least method is named analysis based estimation. This method is based on the fact that in cephalometric analyses the linear and angular relations between landmarks are used and the locations of some landmarks are never used individually. In following, we describe each mechanism and the landmarks that can be detected by that mechanism.

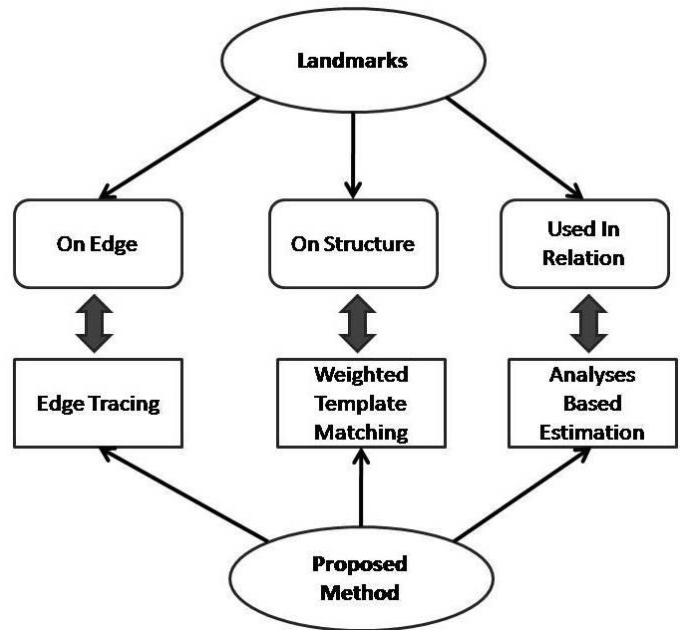

Figure 2. The overall view of proposed method.

It is worth mentioning that to save the time and decreasing the detection error due to image distortion, the search region of each landmark is limited to a certain region where assuredly contains the corresponding landmark and its anatomical characteristics. In this work, the location of certain region of each landmark is determined by a prior knowledge of human skull. To find this region, the mean position of the favorite landmark is measured on training images. The certain region would be a window centered on the mean position of the corresponding landmark and covers the adjacent area which includes the landmarks with a probability of 100%. Figure 3 shows the results of detecting desired certain regions for some landmarks.

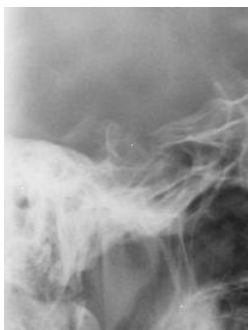
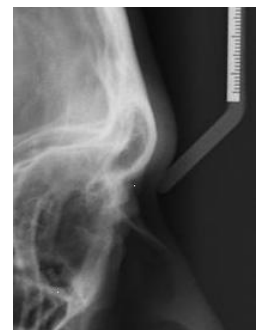

(a)                                                      (b)

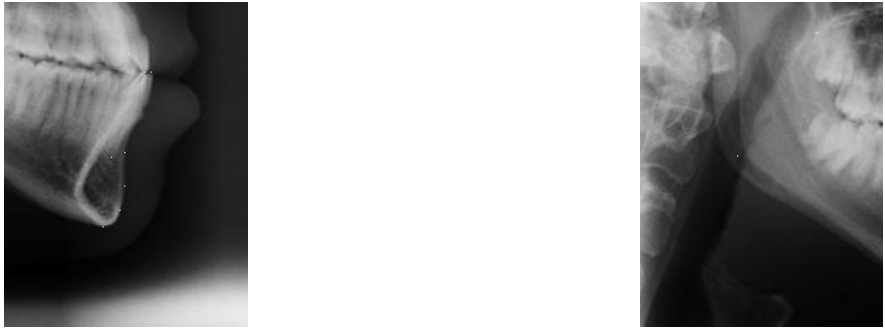

Figure 3. The certain regions for some landmarks. (a) Sella, (b) Nasion, (c) Menton, (d) Gonion.

The next step is to find the landmark location or landmarks relation. As mentioned earlier, this is done by three distinct methods which are described next. A detailed view of the proposed method is shown in Figure 4.

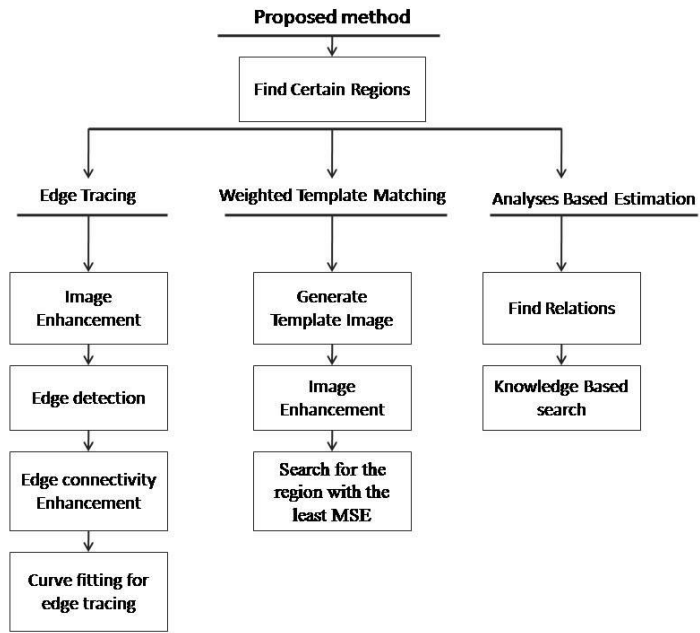

Figure 4. A detailed view of proposed method.

## 2.1. Edge Tracing:

This mechanism is the fastest and most reliable way to predict those landmarks in which according to their anatomical definition they are located on accurately extractable edges. In this mechanism, after finding the certain region of a specific landmark, the edges are extracted and then by tracing a specific edge we can reach the corresponding landmarks located on that edge. It worth mentioning that usually the quality of X-ray images are not sufficiently good, so we use some enhancement method such as adaptive histogram based equalization to aid viewing the key features in the image.

In this mechanism, if a landmark according to its anatomical definition is known to lie on the hard or soft tissue edge of the skull, the Canny edge detector [14] is applied and the extracted edge is traced by the geometrical definition of the landmark. We use Canny edge detection method due to its noise robustness and its ability to extract tin edges.

One of the landmarks that can be efficiently detected by this mechanism is *Menton (Me)* which is defined as the lowest point on the symphyseal shadow of the mandible seen on a lateral cephalogram. In other words it is the lowest point on the bony chin in lateral view. The bony chin can be easily distinguished from other surrounding tissues. Therefore it is possible to extract the edge related to the bony chin and locate the lowest point by following the extracted edge. *Gnathion (Gn)* and *Pogonion (Pog)*, are the other landmarks which could be located by this method. Figure 5 shows the search region for *Pog*, *Me* and *Gn* and the extracted edges in this search region.

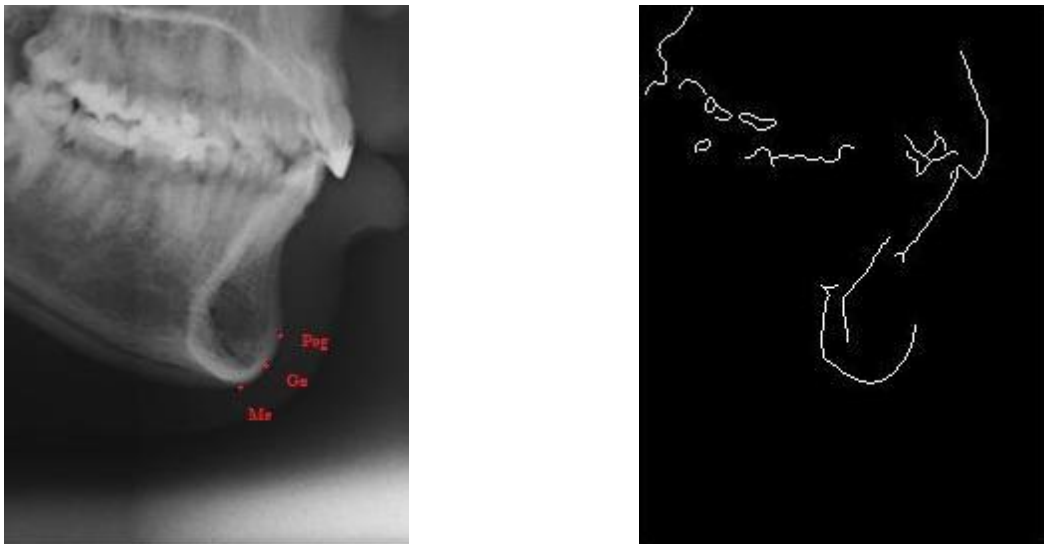

Figure 5. A sample of certain region and its edge detection result.

### 2.2. Weighted Template Matching:

We use Weighted Template Matching (WTM) mechanism to localize those landmarks that are located in an obvious and specific structure which can be extracted or searchable in a given x-ray image. Consequently, according to this specific structure we can define an obvious and unique template image for each of above landmarks belonging to the second category. For example, Sella (S) which is the center of the hypophyseal fossa can be landmarked by this method. As Figure 6 shows, since there is a ladle shaped structure around *Sella* that can be extracted delicately, the best template image for localizing *Sella* is the one that contains this ladle shaped structure.

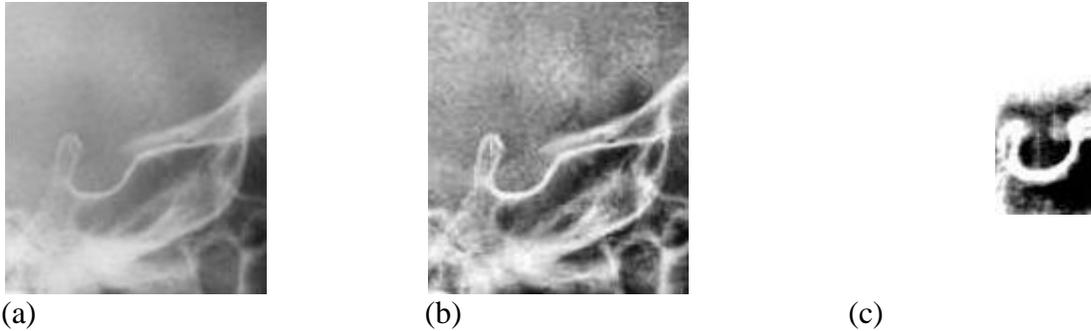

(a)                                      (b)                                      (c)

Figure 6. An example of detected landmark with weighted template matching method. (a) The certain region of Sella. (b) The enhanced region of Sella. (c) The template image for Sella.

There are two steps to apply the WTM mechanism; first, building template image and then searching the best matching of template image in certain region.

The easiest way to have template image for a given landmark is generating a synthetic template image according to the specific structure which are located around the landmark. For example for *Sella* the synthetic template image could be any ladle shaped structure. But it must be considered that this structure is not the same in all cephalograms and differs in size, curvature and hue saturation.

But, this specific structure is not the same in all cephalogram images for example the ladle shaped structure of *Sella* can be different in terms of its size and curvature. So, it is better to average a number of training images where the mentioned landmarks are localized by experts to generate template image instead of artificial generation. From each cephalogram image a region which certainly contains the corresponding landmark and its unique and specific structure are selected and then averaged to obtain the template image. Note that to have a better template image, some enhancement methods are used.

The second step is searching for the template structure through the certain region of the landmark based on MSE (Minimum Square Error) criteria. Since in this method, the template image and the certain region of main image is compared pixel to pixel, these two images are enhanced to ensure that their difference is caused by the structure variety instead of difference in quality of images.

It is noteworthy that all pixels of template image have not the same importance. The pixels that show some features of the favorite structure are more important, therefore we use Weighted Template Matching. To do so the more important pixels must have higher weights in calculating MSE (formula1). The weight of each pixel would be identified manually according to prior knowledge of the specific structures. For example in the ladle shaped structure related to *Sella*, the round shape in lower part of the structure plays an important role in finding template image therefore those pixel would get a higher weight.

$$M \quad (1)$$

**2.3 Estimate based on Cephalometric Analyses Requirements**

Some points such as Orbitale, the lowest point on the inferior rim of the orbit, and Porion, The uppermost point of the external ear meatus, locate in a vague region containing soft issues, nerves and blood vessels. So, detecting such points is too difficult both manually by experts and automatically by intelligent methods. To detect such points, we cannot use two previous mechanisms because they neither have a distinct structure surrounding them which is needed for second mechanism nor are on the extracted edges distinguished by edge detection techniques. In other words, the third category of landmarks contains those that surrounded by vessels and nerves in which make them hard even for experts to find their location and also the mentioned soft tissue makes it hard or even impossible for edge detection techniques to extract the favorite edge.

According to what is mentioned above, one possible way is to use the predefined knowledge of such landmarks and identify the relative location of them based on the location of others, e.g. *Orbitale* is in the almost middle of the line connecting *Nasion* and *Posterior Nasal Spine*, with a usual bias to right and down. This prior knowledge would be helpful to estimate the *Orbitale* more accurately compared with previously mentioned techniques (template matching and edge detection). (Figure 7).

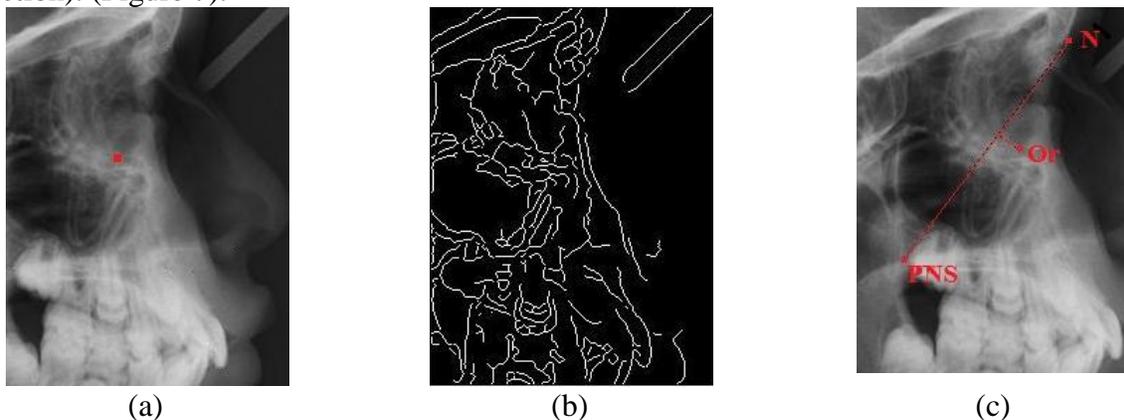

(a)                                            (b)                                            (c)

Figure 7. An example of landmark which is hard to be detected. (a) The Porion landmark and its certain region. (b) the result of edge detection on the Porion's certain region. (c) the relation between the three landmarks: Nasion, Orbitale and PNS.

It must be mentioned that the ultimate aim of landmark location is to do the cephalometric analyses. In cephalometric analyses linear and angular values are produced by connecting the landmarks. The point is that the exact location of some landmarks is just needed to find some indicators which are used in analyses such as the slope of line concatenating two landmarks or the angle between such lines. Consequently, it is possible to find the concatenated line of two landmarks instead of their locations. For example, the Frankfort horizontal plane which is used in most analyses is made by connecting *Orbitale* to *Porion*. Although the location of *Or* could be identified by above methods, the *Po* is really hard to be located because of many similar radiolucencies existing in the certain region of the template matching system that resemble the radiolucency of the internal auditory meatus. (Figure 8(a)) Also no distinguishable edge could be extracted from this certain region.

To overcome this problem, we first measure the inclination of the line connecting *Or* and *Po* which was identified using template matching mechanism. Then, since the line connecting *Nasion* and *Sella* is normally 6-7 degrees above the Frankfort plane, we measure the inclination of this line (S-N) and added 7 degrees to this inclination to find the inclination of the Or-Po line (Figure 8(b)). Then these two obtained inclinations for Or-Po line (S-N+7 and the line connecting automatically founded *Or* and *Po*) were compared with the true horizontal line and the line which had the least different angle with the true horizontal line was accepted as the line of the Frankfort horizontal plane.

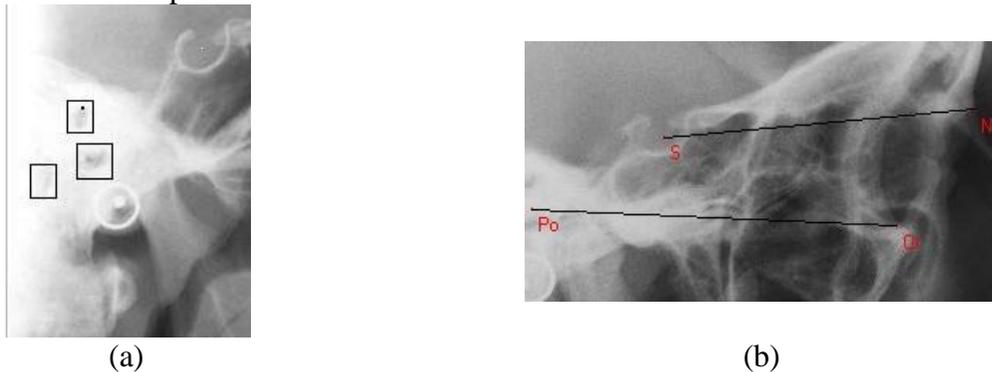

(a)          (b)

Figure 8. (a) many similar radiolucencies existing in the certain region of PO, (b) the connected lines of N-S and Po-Or

In addition, the main aim of detecting the location of *UIAT, UIT, LIA, LIT, Go* and *Me* landmarks are calculating the inclinations of UIA-UIT, LIA-LIT and Go-Me lines. On the other hand, the accurate location of these landmarks was not needed in analyses but the inclination and the angle of these lines was the data required for the analyses. So, here, we estimate these inclinations instead of localizing the above landmarks by following steps: 1) using the edge enhancement technique to find the edges of the image, 2) determining the certain region for the desired line, 3) identifying the best detectable line in the certain region by sampling from the favorite edges, 4) Since the line which was found in step 3 was not a real line but a set of connected dots (pixels), a curve fitting mechanism was used to trace the best fitted line which passed through the dots. Figure 9 shows the obtained UIA-UIT, LIA-LIT and Go-Me lines using such mechanism.

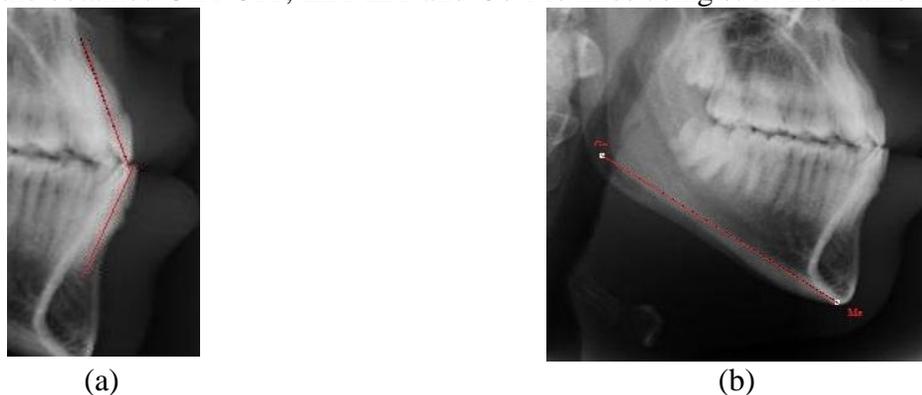

(a)          (b)

Figure 9. (a) The obtained UIA-UIT, LIA-LIT, (b) The Go-Me line

### 3. Experimental Results:

This section presents the experimental results in three subsections. First, the tested datasets are described and second, the performance criteria's and the comparison method is given in detail. The next subsection presents the achieved results of HALD by means of the introduced criteria's are presented. Also the results are compared with some of best known methods in this realm.

### 3.1. Experiments design:

The main aim of this paper is presenting a novel hybrid method for automatic landmark detection in which it is fast and accurate enough to be used as a real word application in the dental medicine centers. Therefore a set of forty digital cephalometric radiographs is randomly selected from the archive of a private oral and craniofacial radiology center to verify the behavior of proposed method. These radiographs have differences in sex, age, racial group, type of occlusion and skeletal pattern. Figure 10 shows some of the selected radiographs.

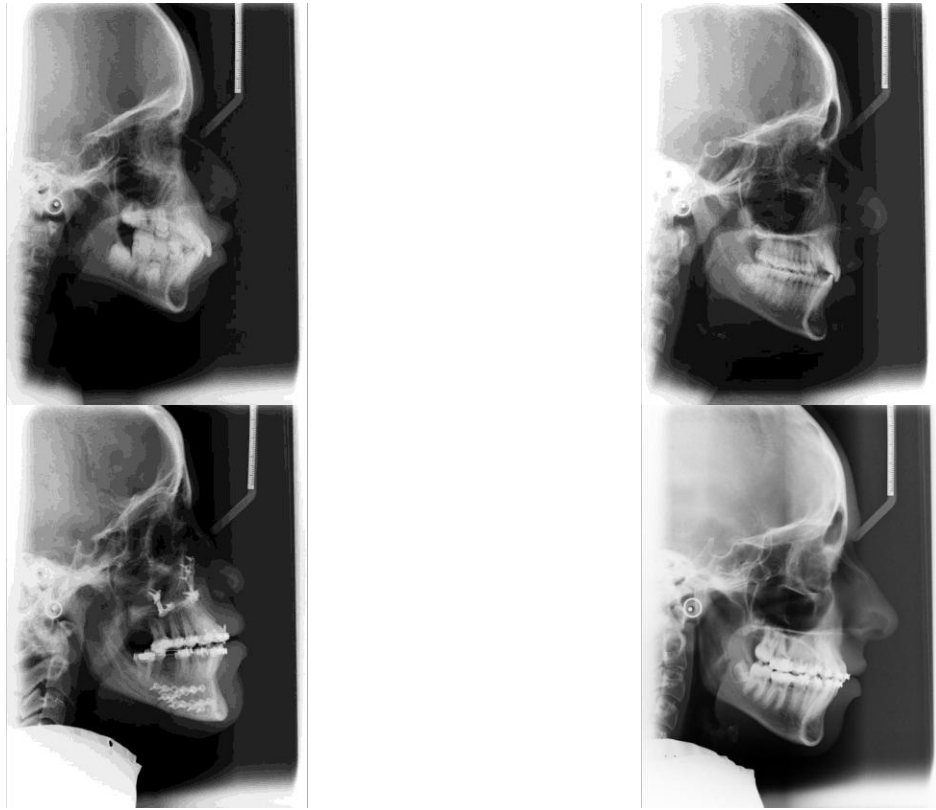

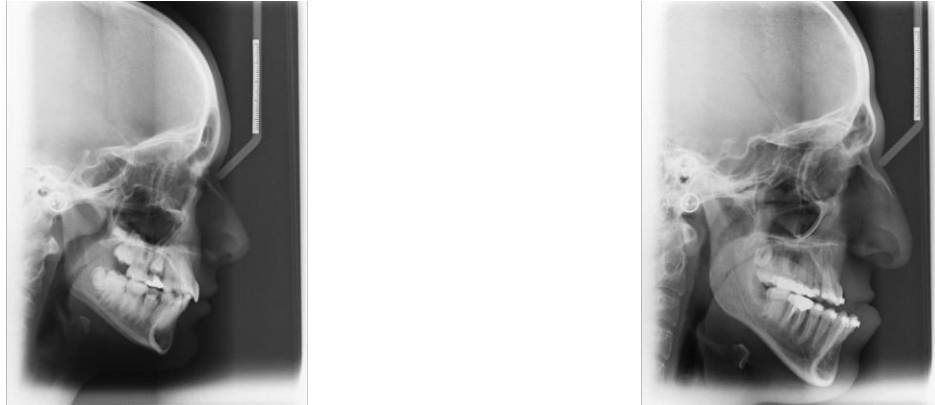
Figure 10. Some of tested cephalograms.

To have a baseline to compare the accuracy of proposed method, the landmarks are manually identified twice by expert orthodontists with at least 6 years of clinical experience. For each landmark, the mean value of these localizations is used as the baseline landmark.

The radiographs are divided into two equal parts as train and test samples for our proposed method. The obtained results by our method were compared with the baseline landmarks and the mean and standard deviation of the correct detection rate is reported.

**3.2. Performance measures:**

To obtain performance of HALD, for each landmark, the average and standard deviation of the algorithm's identification error according to the baseline is calculated. Although there is no standard definition for accuracy in this domain, it is usual to report the ratio of images with mean error less than 4mm, 2mm and 1mm as the accuracy of method for each landmark[]. These values are the common thresholds which shows that the less than 4mm error is acceptable, accurate detection should have less than 2mm error and having less than 1 mm error is ideal.

It is worth mentioning that the above reporting strategy with predefined thresholds is only applicable for the landmarks but not for the inclinations and lines which are detected by the third mechanism. The reason is that the mean errors of the inclinations must be reported in degrees. In order to make it easy for our results to be compared with other studies and also the unity of the reports, we calculated the amount of degrees that was equivalent to the mentioned thresholds of the points at the ends of lines. To obtain the equivalent threshold to x mm predefined threshold for the P1-P2 line, according to Figure 11 the position of each point (P1 and P2) transfer with x/2 mm (if the position of P1 or P2 point can be detected by the first or second mechanism) or x mm amount (if neither P1 and P2 can be independently landmarked). So, the mount of equivalent threshold amount in degrees (α) could be calculated by means of the following formula:

$$\alpha \tag{2}$$

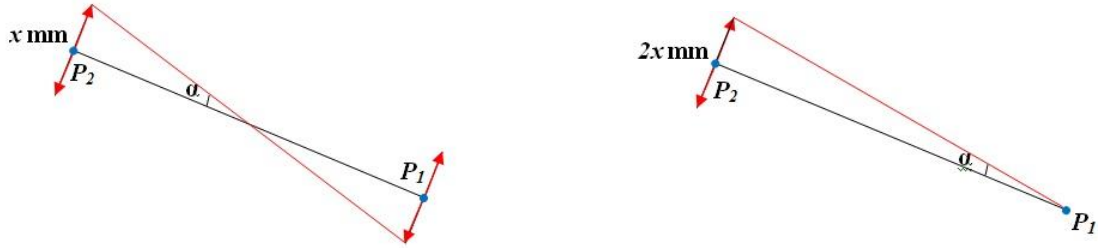

Figure 11. obtain the equivalent threshold to x mm predefined threshold for the P1-P2 line.

Table 2 shows the degrees thresholds approximately equivalent to millimeters thresholds.

Table 2. Degrees thresholds

|  | 1mm | 2mm | 3mm | 4mm |
|---|---|---|---|---|
| **Po-Or** | 0.82 | 1.53 | 2.39 | 3.05 |
| **Go-Me** | 0.89 | 1.68 | 2.61 | 3.35 |
| **UIT_UIA** | 2.23 | 4.34 | 6.61 | 8.64 |
| **LIT_LIA** | 2.41 | 4.7 | 7.15 | 9.33 |

### 3.3. Result and Discussion:

The obtained results of our proposed method, namely HALD, in the mentioned sixteen landmarks are presented in Table 3 in terms of the mean error of detection and its standard deviation. In this study the sixteen landmarks are categorized into three groups: 1) ME, GN and POG, are localized by edge tracing mechanism, 2) A, ANS, B N, Or, PNS, S and UIT are identified by weighted template matching mechanism, 3) Po, Go, UIA, LIA and LIT are determined based on their relation with the other landmarks. In the other words, the four lines Or-Po, Go-Me, UIA- UIT and LIA-LIT are estimated. Table 3 is colored based on these three categories where the white cells show the first group, the gray cells show the second group and the dark gray one show the third group of landmarks.

Table 3. Detection Error of HALD on all landmarks.

| Landmark | Mean ± SD (mm) | Landmark | Mean ± SD (degree) |
|---|---|---|---|
| Me | 0.9±0.6 | Po-Or | 1.9±1.5 |
| Gn | 1.2±0.8 | Or | 2.7±1.4 |
| Pog | 1.4±1.2 | Go-Me | 2.1±1.7 |
| S | 1.4±2.2 | UIT | 1.5±1.5 |
| A | 2.2±1.5 | UIT- UIA | 4.6±3.5 |
| B | 1.4±1.8 | LIT-LIA | 2.7±2.4 |
| N | 1.6±1.1 | | |
| ANS | 2.9±1.7 | | |
| PNS | 2.1±1.8 | | |

On average, HALD has less than 1.7mm error detection for all landmarks and about 2.3 degree error for the inclination of detected lines. The performance of the proposed method according to the introduced thresholds is illustrated in Figure 12. The figure shows the landmarks and lines on the vertical axis. The horizontal axis which is labeled as *performance* shows the ratio of test images where the detection error is less than the defined threshold. Also these results are shown in Table 4 in percentages.

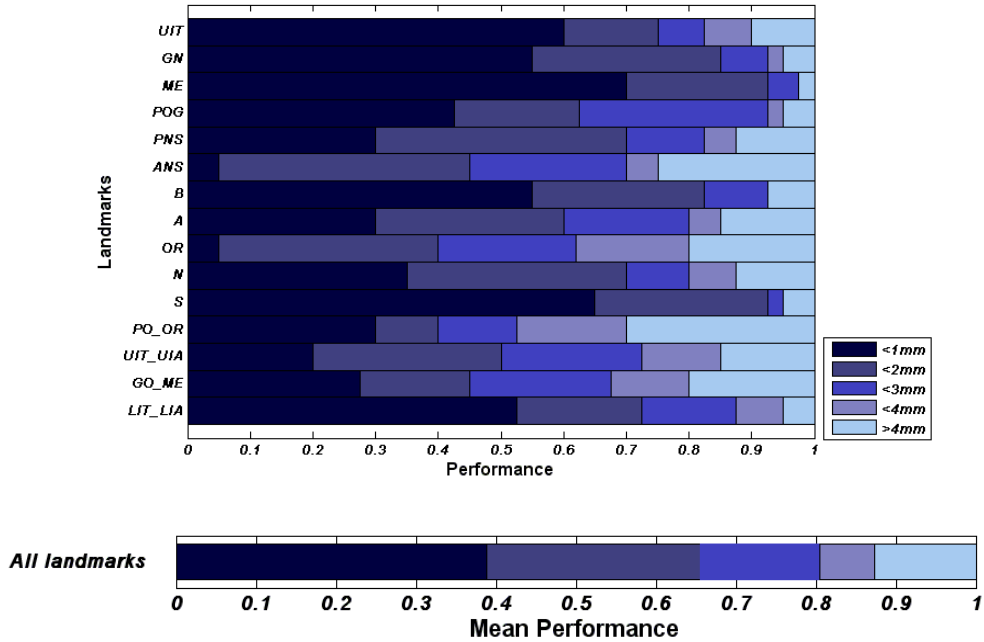

Figure 12. The performance of HALD. The ratio of test images where the detection error is less than the defined threshold is defined as performance in each landmark.

Table 4. The performance of HALD in percentages.

|        | < 1mm | < 2mm | < 3mm | < 4mm |
|--------|-------|-------|-------|-------|
| UIT    | 60%   | 75%   | 82.5% | 90%   |
| GN     | 75%   | 85%   | 92.5% | 95%   |
| ME     | 70%   | 92.5% | 97.5% | 97.5% |
| POG    | 42.5% | 62.5% | 92.5% | 95%   |
| PNS    | 30%   | 70%   | 82.5% | 87.5% |
| ANS    | 5%    | 45%   | 70%   | 75%   |
| B      | 55%   | 82.5% | 92.5% | 92.5% |
| A      | 30%   | 60%   | 80%   | 85%   |
| OR     | 5%    | 40%   | 62%   | 80%   |
| N      | 35%   | 70%   | 80%   | 87.5% |
| S      | 65%   | 92.5% | 95%   | 95%   |
| PO-OR  | 30%   | 40%   | 52.5% | 70%   |
| UIT-UIA| 20%   | 50%   | 72.5% | 85%   |
| GO-ME  | 27.5% | 45%   | 67.5% | 80%   |
| LIT-LIA| 52.5% | 72.5% | 87.5% | 95%   |

| All landmarks | 38.83% | 65.50% | 80.47% | 87.34% |

Figure 13 shows the comparison of our method with some well known approaches in this realm. Parthasarathy et al. [2], Rudolph et al [3] Tong et al [15] used the Knowledge based approaches by image processing techniques. From model based approaches, two best known methods are Saad et al [6] and Hutton et al [5]. Grau et al [12] has the best results among other soft computing techniques. Finally Liu et al. [13] and Giordani et al. [8] are chosen among hybrid techniques.

The figure shows the name of methods according to their authors on the vertical axis and displays the performance on the horizontal axis. Here, for each method, the performance is measured as the mean of performances on all the landmarks. As it is shown in the figure, HALD outperforms most of the other techniques except *Grau* [9] which uses neural networks as its base method.

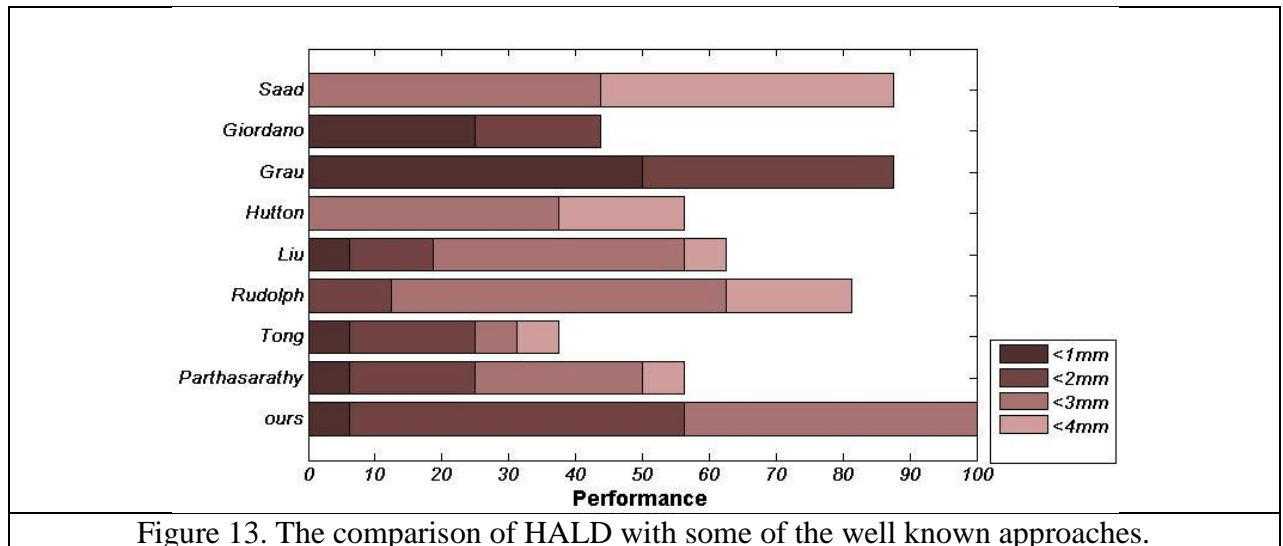

Figure 13. The comparison of HALD with some of the well known approaches.

## 4. Conclusion

Here, automatic landmark detection is investigated and regarding the nature of landmarks, it is described that a single method cannot find all the landmarks effectively. While image processing techniques are the best answer for estimating the location of some landmarks, it is not applicable for the others. There is the same issue in applying the other single approaches such as model based and soft computing techniques. In this work a new hybrid approach called HALD is presented where three different detection methods are combined to detect the landmarks according to their nature. HLAD tries to localize sixteen landmarks which are used in the best known cephalometric analyses. These landmarks are grouped in three categories based on their anatomical features. The first category of landmarks contains the ones that are located on edge and detected using Edge tracing. The landmarks of the second category where a unique structure is found around the landmark are detected using weighted template matching technique. The last but not the least category consists of the other landmarks which do not belong to either of those categories. In the point of fact, we propose an analysis based approach to detect the landmarks of third category, since whilst the exact position of some landmarks is directly used in the analyses, there are several landmarks which their relation with the other landmarks are used instead.

Regarding to the reported results, HALD showed a promising performance and consequently the proposed approach seems to be practicable in the laboratories since its overall estimation error is acceptable. Since in each research some special landmarks is detected and there is no standard list of needed landmarks, it is not easy to compare our method with others. However, the comparison results show that HALD can outperform other methods on average because it is a hybrid approach and after categorizing the landmarks suggests the best way to detect the landmarks of each category.